%% file: main.tex
\documentclass[letterpaper, 10 pt, conference]{ieeeconf} 

\IEEEoverridecommandlockouts                              % This command is only needed if 
\usepackage{graphics} % for pdf, bitmapped graphics files
\usepackage{epsfig} % for postscript graphics files

\usepackage{xspace}
\usepackage{amsmath}
\usepackage{amssymb}
\usepackage{mathtools}
\usepackage{cite}

\usepackage{xcolor}
\usepackage[normalem]{ulem}

\usepackage{color}
\usepackage{booktabs}
\usepackage{url}

\usepackage{graphicx}

\usepackage{multirow}
\usepackage{anyfontsize}

\usepackage{hyphenat}
\usepackage{hyperref}
\usepackage{subcaption}

\usepackage{xspace}

\input{usefulCommands}

\usepackage{dsfont} % for sets like the reals

\title{\LARGE \bf
Validation of Simulation-Based Testing: Bypassing Domain Shift\\
with Label-to-Image Synthesis
}

\author{Julia Rosenzweig$^{* \dagger \wedge} $  \and Eduardo Brito$^{* \dagger \wedge}$  \and Hans-Ulrich Kobialka$^{\dagger}$ \and Maram Akila$^{\dagger}$    \and Nico M. Schmidt$^{\mathparagraph}$ \and Peter Schlicht$^{\mathparagraph}$ \and Jan David Schneider$^{\mathsection}$ \and Fabian Hüger$^{\mathparagraph}$ \and Matthias Rottmann$^\ddagger$ \and Sebastian Houben$^{\dagger}$ \and Tim Wirtz$^{\dagger \wedge} $   % $^{*}$% <-this % stops a space
\date{February 2020}

\thanks{$*$ These co-first authors contributed equally to this work and are the corresponding authors: 
{\tt\footnotesize $\{$julia.rosenzweig, eduardo.alfredo.brito.chacon$\}$@iais.fraunhofer.de} }
\thanks{$\dagger$ The authors are with the Fraunhofer Institute for Intelligent Analysis and Information Systems}
\thanks{$\wedge$ These authors are with Fraunhofer Center for Machine Learning, Sankt Augustin, Germany}
\thanks{$\mathparagraph$ These authors are with CARIAD SE}
\thanks{$\mathsection$ The author is with Volkswagen AG}

\thanks{$\ddagger$ The author is with the University of Wuppertal, Dept.\ of Mathematics}
}

\begin{document}

\maketitle
\thispagestyle{empty}
\pagestyle{empty}

%%%%%%%%%%%%%%%%%%%%%%%%%%%%%%%%%%%%%%%%%%%%%%%%%%%%%%%%%%%%%%%%%%%%%%%%%%%%%%%%
\begin{abstract}
Many machine learning applications can benefit from simulated data for systematic validation - in particular if real-life data is difficult to obtain or annotate. However, since simulations are prone to domain shift \wrt real-life data, it is crucial to verify the transferability of the obtained results.
\par
We propose a novel framework consisting of a generative label-to-image synthesis model together with different transferability measures to inspect to what extent we can transfer testing results of semantic segmentation models from synthetic data to equivalent real-life data. With slight modifications, our approach is extendable to, \eg general multi-class classification tasks. Grounded on the transferability analysis, our approach additionally allows for extensive testing by incorporating controlled simulations. We validate our approach empirically on a semantic segmentation task on driving scenes. 
Transferability is tested using correlation analysis of IoU and a learned discriminator.
Although the latter can distinguish between real-life and synthetic tests, in the former we observe surprisingly strong correlations of 0.7 for both cars and pedestrians.
\end{abstract}

\input{1_intro}
\input{2_related_work}
\input{3_approach}
\input{4_experiments}
\input{5_conclusion}

\vspace{0.3cm}\noindent

\section*{Acknowledgments}
This work will be presented at the 4th Workshop on “Ensuring and Validating Safety for Automated Vehicles” (WS13), IV2021.\\
\par The work of Hans-Ulrich Kobialka, Maram Akila and Sebastian Houben was funded by the German Federal Ministry of Education and Research, ML2R - no. 01S18038B.

\bibliographystyle{IEEEtran}
\bibliography{references}

\end{document}

%% file: usefulCommands.tex
\newcommand{\ie}{i.e.,\ }
\newcommand{\eg}{e.g.,\ }
\newcommand{\wrt}{w.r.t.\xspace}
\newcommand{\cf}{cf.\xspace}
\newcommand{\etal}{et al.\xspace}

\definecolor{negativered}{RGB}{128, 0, 0}
\definecolor{positivegreen}{RGB}{0, 128, 0}

\definecolor{blue1}{RGB}{53, 113, 152}
\definecolor{blue2}{RGB}{65, 141, 191}
\definecolor{blue3}{RGB}{102, 164, 205}
\definecolor{blue4}{RGB}{139, 187, 218}

\newcommand{\reffig}[1]{Fig.~\ref{#1}}

\newcommand{\refsec}[1]{Sec.~\ref{#1}}

%% file: 1_intro.tex
\section{Introduction}

\noindent
With an ever-increasing prevalence of machine learning (ML) models in many industrial sectors including safety-critical domains like autonomous driving (AD), medicine and finance, the topic of testing ML models is currently an active field of research. This is mainly driven from the need to verify, safeguard and certify such models, \cf  \cite{BRAIEK2020110542, MLtestingsurvey2020}. 
In this context, many deep learning models demand enormous amounts of labeled representative data for training but even more so for systematic testing and validation. 
Often, single datasets do not provide the necessary data diversity and statistically assessing the performance of a model in all situations of interest is not possible (since \eg near-accidents in real scenes, fortunately, occur rather rarely). 
As a result, testing on real data can be prohibitively expensive or even unfeasible: Some classification tasks on video for AD involve recording and manually labeling hours of street scenes or finding rare critical situations in actual driving data. This motivates training and testing models with data from simulations \cite{simulation_based_testing20, wagner19,CARLA2017}, which can be seen as a particular approach within informed ML, \cite{vonrueden2020informed, von2020combining}. 
\par Synthetic data generation essentially comes ``for free'': 
Virtual environments allow for fast and controllable generation of scenes and thereby enable higher (test) coverage and systematic statistical testing,
\eg against occlusions and corner cases, which is unfeasible in real scenes.
However, using synthetic data introduces a domain shift with respect to the real data that the model was trained on. 
In this work, we focus on the question of reliable and realistic testing of semantic segmentation models with synthetic data, namely:
\begin{itemize}
\item[] \emph{To what extent can we transfer testing results from synthetic data to real data?}
\end{itemize}
\noindent
In particular, we investigate if testing on synthetic data uncovers exactly the same failure modes that would occur in a real environment.
We propose to tackle this issue with a modular two-stage framework that allows for an in-depth investigation beyond aggregated performance scores.
It combines label-to-image synthesis with controllable simulations enabling the generation of test cases, e.g.\ a child jumping on the street or any situation stated in safety requirements of regulatory entities.
In the first step, a paired set of semantically equivalent real and synthetic scenes allowing a direct comparison (of \eg model performance) is used to assess the transferability of testing results.
If it is satisfactory,\footnote{For the approach to be useful results do not need to be identical. However, a strong correlation in, \eg failure modes or performances for investigated classes is desirable. Due to the pairwise correspondence such quantities are directly measurable.} in the second step test cases can be generated independently from the previous set using, \eg a simulator to obtain labels for scene synthesis.
In this way, the testing process ``bypasses'' the domain shift.
With the perspective of avoiding repeated costly tests on real data, various participants involved in testing ML models can benefit from this workflow, be it as a developer or  auditor in certification bodies.

This work is organized as follows: \refsec{sec:related_work} outlines some topics related to our work. \refsec{sec:approach} is dedicated to the description of our conceptual framework, including the data generation procedure as well as the validation measures. We validate our framework in \refsec{sec:experiments} for the use case of semantic segmentation for AD as a proof of concept. Finally, we conclude in \refsec{sec:conclusion} discussing some open questions and giving an outlook on future work.

%% file: 2_related_work.tex
\section{Related work}
\label{sec:related_work}

We outline some directions of related work. One concerns the testing of ML models, others are domain adaptation and synthetic data generation.

\subsection{Testing of Machine Learning Models}
\label{sec:related_work_testing}
Our approach falls into the broader category of offline testing methods (in contrast to online testing methods,
which are applied at deployment time), \cite{online_offline_testing}, and aims to find and test weak spots. By doing so, it opens up possibilities for simulation-based \emph{safety argumentations} \cite{safety_args2018, safety_args_2020_schwalbe}.
Often the conventional software-testing approaches are not directly applicable for testing ML-based systems or need strong adaptations to be applied.
Hence, new \emph{methods, measures, and evaluation techniques} are needed, \cf \cite{BRAIEK2020110542, MLtestingsurvey2020}, 
to argue for the safety of ML models. 
In \emph{statistical model checking} \cite{barbier2019validation}, ML models are validated using Key Performance Indicators of interest in combination with statistics and modeling the ML components as probabilistic systems. 
In this paper and in many other testing approaches, \emph{simulation-based testing}, \eg \cite{CARLA2017,simulation_based_testing20,wagner19}, is deployed in which simulators are used to create testing data (\cf \refsec{sec:related_work_domains}).
However, the frameworks proposed in this context  in  part or entirely neglect validating their results in real-world situations.
Our approach addresses exactly this shortcoming by proposing a framework to assess to what degree testing results obtained on synthetic data are realistic.
Closest to our contribution is the work by Wagner \etal \cite{wagner19}. The authors propose an approach to locally verify the use of simulation data for testing. Our approach differs significantly in two main aspects: First, we do not apply any formalism or scene description language to produce corresponding synthetic data from the real data. Instead, we  use the real labels as input to a generative model to produce a paired synthetic dataset (see \refsec{sec:data_generation}). Additionally, due to the use of the generative model (also on simulation labels, see \refsec{sec:approach_extended_dataset}), we can extend the input space coverage not only locally and allow for completely new (controllable) scenarios to be generated and tested against.
This is in line with the literature about \emph{scenario-based testing} in the context of AD \cite{Neurohr20, Bussler20}.

\subsection{Domain Adaptation and Synthetic Data Generation}
\label{sec:related_work_domains}
While simulations may help with the aforementioned challenges, synthesizing data for testing creates a domain mismatch \wrt the data the system will process when deployed in real-world settings. 
Most \emph{domain adaptation} approaches aim to measure and minimize the domain gap between a source and the desired target domain during the training phase so that the system generalizes well on the target domain  \cite{kouw2019review,Mei2018}. 
In contrast, we aim to validate transferability of testing results on synthetic data to real-world data regardless of the concrete magnitude of the domain gap.
To this end, we apply a generative adversarial label-to-image synthesis model. 
In the particular case of video data employed for our proof of concept,  \emph{video-to-video synthesis}, \cf \cite{Chen_2017_ICCV, vid2vid}, can serve as a generative model to synthesize photo-realistic time-consistent videos given an input video. 
This task can be defined as a distribution matching problem that can be solved by conditional generative adversarial models. 
It can also be understood as an \emph{image-to-image translation} problem, \cf  \cite{Isola_2017_CVPR, pix2pix_follow_up}, additionally introducing the necessary temporal dynamics so that the synthesized image sequences are temporally coherent.
\par Other approaches try to directly learn generative models for synthesizing realistic scenes that have a low domain gap compared to the real data, \cf \cite{Meta-Sim, Meta-Sim2}, or learn models that improve the realism of given simulated images, \cf \cite{Shrivastava_2017_CVPR}. 
Close to the idea for our data processing procedure is the work \cite{Virtual_Kitty}, where the authors create a simulated world that is cloned from the corresponding real-world dataset \cite{Kitty}. \\

%% file: 3_approach.tex
\section{Approach}
\label{sec:approach}
Our ultimate goal is to assess whether the output, which the system under test produces on synthetic data is the same as the one produced on corresponding real-life data.
Here, the correspondence is defined through representations of the same abstract events in the real and the synthetic domain. For instance, the abstract scenario of driving near a crosswalk with pedestrians has representations in both the real, \eg an image recorded directly on the street, and synthetic domain. 
Without domain gap, testing the network on both representations should ideally lead to exactly the same output. 
As, however, domain gaps can occur in practice, we propose a qualitative framework to validate the transferability of the testing results by exploring this gap under controlled circumstances.
Although we detail it for semantic segmentation, it is also applicable for other multi-class classification tasks with slight modifications.
\par
Our framework consists of the following basic components:
\begin{enumerate}    
    \item An ML model (system under test) trained for a specific semantic segmentation task on real data featuring the target appearance\footnote{We use the expression ``target appearance'' to describe the characteristics of the real-world data in distinction to its synthetic counterpart.}, 
    \item a labeled real-world testing dataset featuring the target appearance,
    \item a generative label-to-image transfer model, which synthesizes data featuring the target appearance from segmentation masks (or from classification labels with sufficient semantic information, \eg from meta data) 
    
    \item a controllable simulation engine that allows us to create labeled synthetic data of interest, and
    \item a set of (interpretable) testing measures to evaluate transferability of testing results on synthetic data to corresponding real data. 
\end{enumerate}
We initially apply testing measures to validate that the model behavior on synthesized data obtained from the generative model is indicative of its behavior on real data, thereby exploring the domain gap.
If our testing measures yield a robust correlation of results between synthetic and real data, this substantiates the transferability of results and reliability of testing\footnote{Note that \wrt testing this can only be seen as a qualitative indicator of transferability.}.
In this case, testing on simulated scenes of interest, processed by the generative model, can be expected to yield valid results.
The intended validation of transferability (described in more detail in \refsec{sec:approach_validation}) requires a specific data generation procedure (described in \refsec{sec:approach_datasets}). 

\subsection{Data Generation}
\label{sec:approach_datasets}
First, we describe how to apply the generative model to produce pairs of real and synthesized elements to validate transferability. Secondly, assuming the transferability of results from synthetic to real data is valid, we detail the extension of the input data coverage by generating additional data with our controllable simulation engine involving the same generative transfer procedure.
Here, we have the following underlying assumption (shown in \reffig{fig:approach_data_generation}): Focusing on the ground truths as input for the generative model implicitly makes use of the advantage that the per-element domain gap between the ground truth's of simulated and real data is smaller than between their respective input data. Thus, the synthesized elements resulting from simulation masks and those resulting from real masks exhibit a smaller domain shift enabling argumentation of transferability. This way, we shift the questions concerning the relevance of the domain gap from the simulation to a controllable comparison.

\subsubsection{Real-Synthetic Paired Dataset}

\label{sec:data_generation}
\begin{figure}[t]
\centering
        \includegraphics[width=0.99\columnwidth]{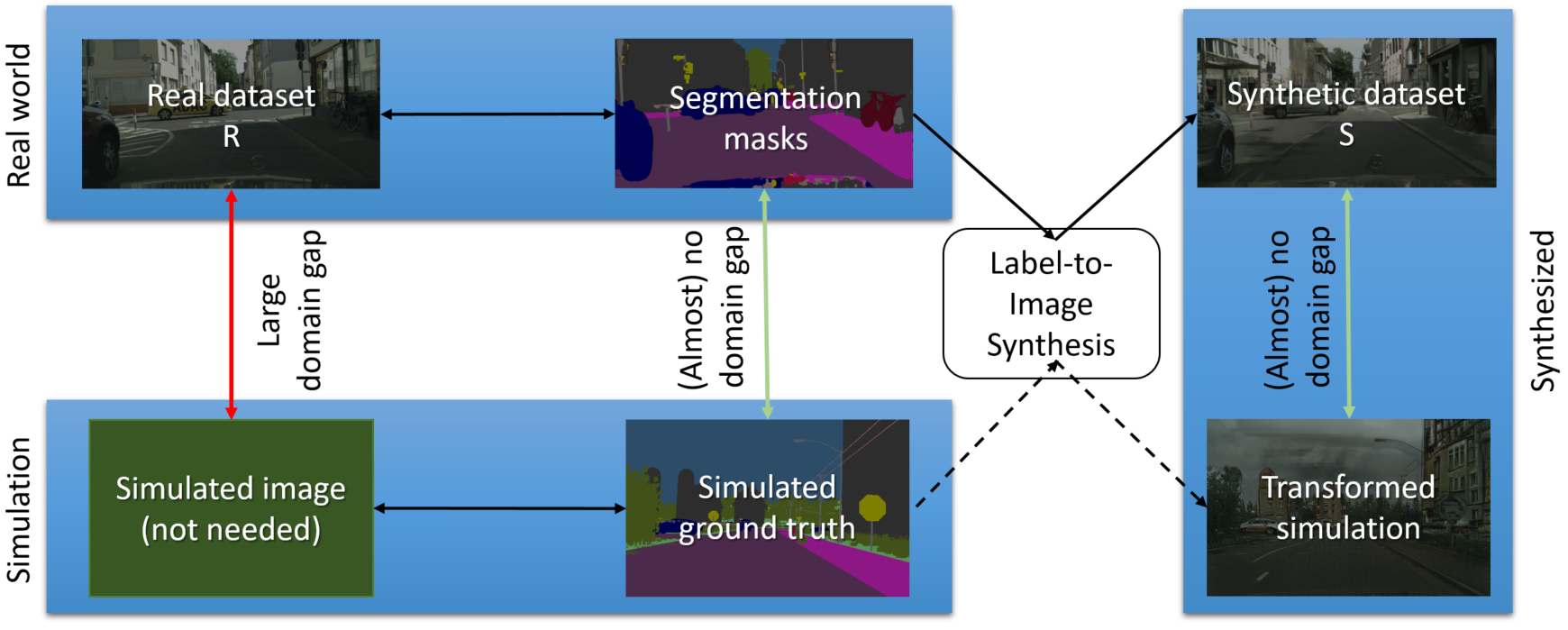}
\caption{Label-to-image-based synthesis: Generation of the synthetic counterpart to the real set (upper part) and transformation of the simulated ground truth masks (lower part)}
\label{fig:approach_data_generation}
\end{figure}

We first need a paired dataset to validate transferability. 
Let $R$ be our initial, labeled dataset extracted directly from the real-world data that features the target appearance.  
We generate a corresponding synthetic dataset $S$ in the following way: For each element $r_i\in R$, we synthesize an equivalent element $s_i\in S$, whose label 
is the same as $r_i$, by applying the generative model to the segmentation label of the real element with the intention of mimicking the target appearance of $R$.
The procedure is depicted in the upper part of \reffig{fig:approach_data_generation}. 
In case the labeled elements from $R$ do not suffice to evaluate transferability and additional unlabeled elements featuring the target appearance are available, $R$ can be extended with pseudo-labels and $S$ with the respective synthesized elements.

\subsubsection{Extending Input Coverage: Transformed Simulated Dataset} 
\label{sec:approach_extended_dataset}
Corner cases and rare events do not appear frequently enough in many datasets directly extracted from the real world to allow for extensive testing.
However, many use cases including safety-critical ones require sufficient test coverage and a controllable framework can help to extend the testing.
For this purpose, the simulation engine synthesizes labels containing cases of interest\footnote{Note that for this procedure, we don't need the actual simulated elements but only their labels.}. 
We then apply the same generative model as in \refsec{sec:data_generation} so that the synthesized data has the same style as $R$, see the lower part of \reffig{fig:approach_data_generation}. Note that here the labels have to be in the same format as used for the paired dataset. 

\subsection{Validation of transferability}
\label{sec:approach_validation}
\subsubsection{Correlation and performance analysis}
\label{sec:approach_validation_correlation}
To assess transferability, we measure the performance of the model on all elements $r_i$ of the real dataset $R$ and on all corresponding elements $s_i$ of the synthesized, paired synthetic set S (\cf \refsec{sec:approach_datasets}). 
We obtain two sequences $(\mathrm{perf}_{r_i})_{i=1}^{|R|}$ and $(\mathrm{perf}_{s_i})_{i=1}^{|R|}$ of performance scores of same length. As a first step, we
compute the sample correlation coefficient of $(\mathrm{perf}_{r_i})_{i=1}^{|R|}$ and $(\mathrm{perf}_{s_i})_{i=1}^{|R|}$.  
The higher the correlation coefficient, the stronger the evidence for transferability of overall qualitative model behavior from the synthetic to the real-world dataset. 
A high correlation coefficient on the paired set and the fact that the transformed extended input scenes have (almost) no domain gap \wrt the  synthetic paired set justifies further testing with the transformed simulated dataset (\cf \refsec{sec:approach_datasets}). 
In a second step, we assess the model performance on it using the same performance measure, $\mathrm{perf}$, as in step one. We consider the aggregation of the per-image scores into one global score as a first proxy for potential performance on corresponding real data. 

\par
However, prediction errors can cancel out in case of non-binary performance scores, like mean-intersection-over-union (mIoU)
, or add up to the same global score even if they are of different nature (\eg in different regions of an image to be segmented).
Hence, global model performance measures, such as the mentioned aggregated score, cannot reflect the complete model behavior to inspect our main questions. Therefore, we need a more differentiated analysis considering error distributions.

\subsubsection{Error distribution analysis}
\label{sec:approach_validation_radarplots}
So far, the approach was application-agnostic. 
For simplicity of notation, we here assume a semantic segmentation task. With slight modifications, the below approach can be adjusted to \eg other multi-class classification-related tasks.
We first introduce some notation to formalize our approach for the error distribution analysis. Let the semantic segmentation problem have $C \in \mathbb{N}$ classes,
let $o(x_i)$ be the prediction mask of the segmentation network to the input  $x_i \in \{r_i,s_i\},~ r_i\in R,~s_i\in S,$ and denote by $y_i$ the corresponding segmentation ground truth mask (the mask is the same for both paired elements). 
Further, let \mbox{$Y_c(i) \coloneqq \{y_i = c\}$} be the subset of $y_i$ that has class $c$, 
for $c\in \left\{1, \dots C\right\}$ (\ie pixels of an image that belong to that class).  
Let \mbox{$O_{c,k}(i) \coloneqq \{e \in Y_c(i) : o(x_i) =k\}$} be the subset of $Y_c(i)$ in which the model outputs class $k \in \left\{1, \dots C\right\}$.
Now, in a second part of our transferability assessment, we analyze the error distribution  on the real dataset $R$  compared to its corresponding synthesized 
set $S$. We construct a confusion matrix per class of interest. That is, per fixed class $c \in \{1, \dots, C\}$ and element $x_i$, we save the true positives as well as false negatives - distinguishing w.r.t. all other classes -  and normalize these values so that the resulting values add up to one: $\mathrm{TP_{c}}(i) = \frac{|O_{c,c}(i)|}{| Y_c(i)|}$ and \mbox{ $\mathrm{FN_{c,k}}(i) = \frac{|O_{c,k}(i)|}{| Y_c(i) |}$ for $k \neq c$}.
Finally, we average over all elements of the respective dataset, resulting in one relative mean true positive score \mbox{$\mathrm{TPS_c} \coloneqq \frac{1}{|R|} \sum_{i=1}^{|R|} \mathrm{TP_{c}}(i)$} and $C-1$ false negative scores \mbox{$\mathrm{FNS_{c,k}} \coloneqq   \frac{1}{|R|} \sum_{i=1}^{|R|} \mathrm{FN_{c,k}}(i)$} for $k\neq c$ \wrt the ground truth class of choice $c$. Note that a $\mathrm{TPS_c}$ score of one is ideal.
This procedure is repeated for all classes $c$. The resulting quantities $\mathrm{TPS_c}$ and $\mathrm{FNS_{c,k}}$, $c\in \{1, \dots, C\}$, now are compared per class across the real and synthetic datasets resulting in a detailed analysis about whether the same misclassifications are made. 
This in turn provides evidence as to how far qualitative mistakes and semantic failures identified in one of the datasets also constitute errors on the other dataset.
Comparing these findings from the paired set with the error distributions on the transformed simulated dataset provides an additional plausibility check. 
It aims to substantiate that testing on the transformed simulated scenes is justified and a corresponding real scene would lead to comparable model behavior. However, since the extended set might contain intentionally challenging elements, the error distribution may deviate. This points to semantic concepts that could lead to failure modes in corresponding real data.
\par We propose to visualize the findings with radar plots, one for each class $c$, since they allow for a systematic visual comparison and readable overview of errors on the different datasets. The axes in the plots correspond to the classes and to guide the eye lines are connected, 
the values being $\mathrm{TPS_c}$ for the ground truth class $c$ and $\mathrm{FNS_{c,k}}$ for the other axes $k\neq c$, \cf \reffig{fig:radar_plots} for an example. 
In addition, we propose boxplots of the distributions of the errors across the elements of the datasets, where each boxplot has the perspective of one ground truth class and shows the distributions of the $\mathrm{TP_{c}}(i)$ and $\mathrm{FN_{c,k}}(i)$  scores \wrt $c$.

\subsubsection{Discriminating Model Outputs and Errors}
While it provides a more comprehensive analysis than solely comparing aggregated performance scores, our error distribution analysis still lacks detail about where exactly these errors happen: 
we cannot assess whether the model behavior (especially regarding the kinds of committed errors as \eg errors in different regions of a segmentation prediction) 
on the synthetic dataset is indistinguishable from the behavior on the real-world dataset. 
By training a discriminator to distinguish between model output on real data and synthesized data (or model errors, respectively) we target exactly this question. The underlying assumption is that if a discriminator cannot distinguish between model outputs/errors to real and synthetic input, they are sufficiently close in the sense that we can consider the model behavior to be "the same" (up to discriminator precision) and, thus, synthetic testing 
is realistic. By choosing an interpretable discriminator setup, we can enhance the interpretability of the differences in model behavior on the real and synthetic sets.
 

%% file: 4_experiments.tex
\section{Experimental Setting}
\label{sec:experiments}
    We validate our approach described in \refsec{sec:approach} for an AD scenario by assessing transferability of tests on simulation videos. 
    Our setup consists of the following components:
\begin{itemize}
    \item The HRNetV2-W48 neural network \cite{HRNet2020}\footnote{\hfill \url{https://github.com/HRNet/HRNet-Semantic-Segmentation/tree/pytorch-v1.1} } for semantic segmentation with $19$ classes trained on the Cityscapes training set \cite{Cordts2016Cityscapes}. We use without further modification the pretrained version 
    available for download\footnote{ \url{ https://onedrive.live.com/?authkey=\%21AErsWO7\%2DxcLEVS0&cid=F7FD0B7F26543CEB&id=F7FD0B7F26543CEB\%21169&parId=F7FD0B7F26543CEB\%21166&action=locate} },
    \item the Cityscapes high resolution (2048 x 1024) validation set serving as the testing dataset,
    \item the generative adversarial video-to-video synthesis tool vid2vid \cite{vid2vid}\footnote{ \url{https://github.com/NVIDIA/vid2vid}}, in particular, without further modifications the variant pretrained on Cityscapes,
    \item the controllable CARLA simulation engine \cite{CARLA2017}, and
    \item as the testing measures, we employ the score mIoU
    as the performance measure for the correlation analysis, an error distribution analysis for each of the  $19$ Cityscapes classes and SkopeRules as rule learner based on the feature engineering from MetaSeg \cite{MetaSeg}\footnote{We use an internal code base provided by the authors of \cite{MetaSeg} which is an extension to their existing repository under \url{https://github.com/mrottmann/MetaSeg}} as discriminator of model outputs and errors.
\end{itemize} 

The technical implementation is detailed further in the following subsections.

\subsection{Generating synthetic data}
\begin{figure}[t]
\centering
\includegraphics[width=0.45\textwidth]{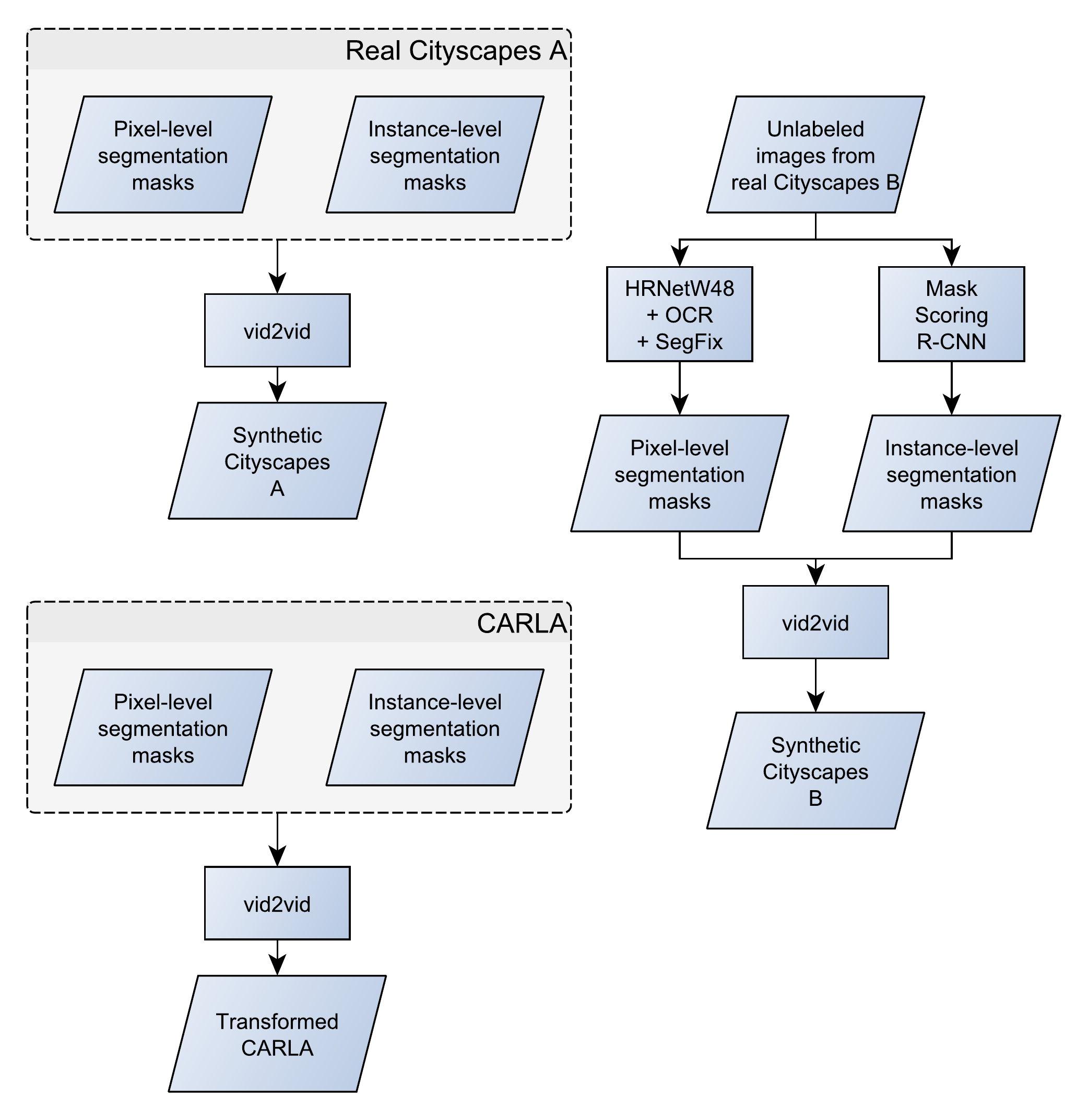}
\caption{Depiction of the performed synthetic generation processes with the involved datasets.}
\label{fig:experiments_data_generation}
\end{figure}

\begin{figure*}[t]
\centering
    \begin{subfigure}[l]{0.32\textwidth}
        \includegraphics[width=\textwidth]{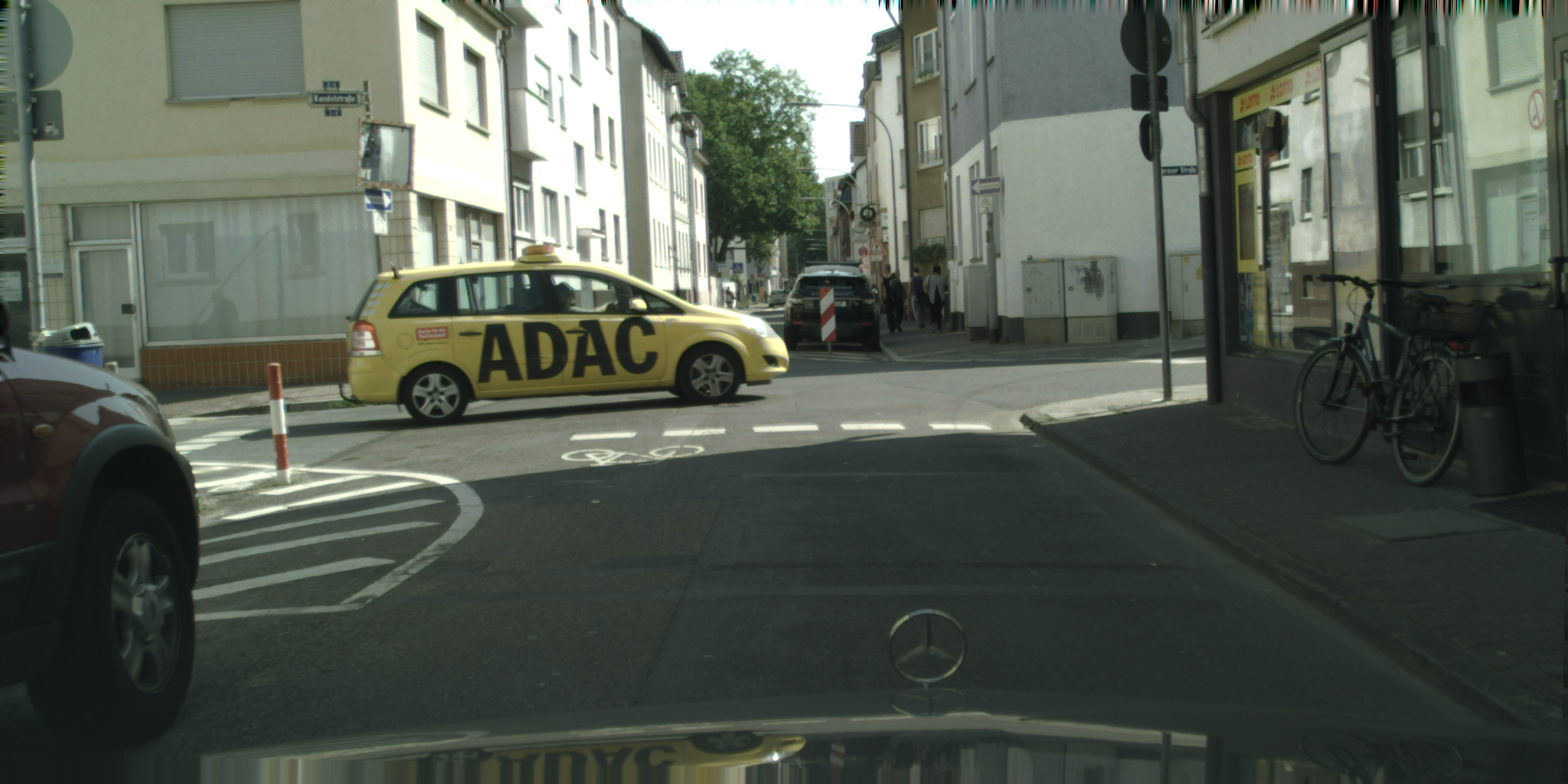}
        \caption{Share of Cityscapes B}
        \label{fig:original_cityscapes}
    \end{subfigure}
    \begin{subfigure}[c]{0.32\textwidth}
        \includegraphics[width=\textwidth]{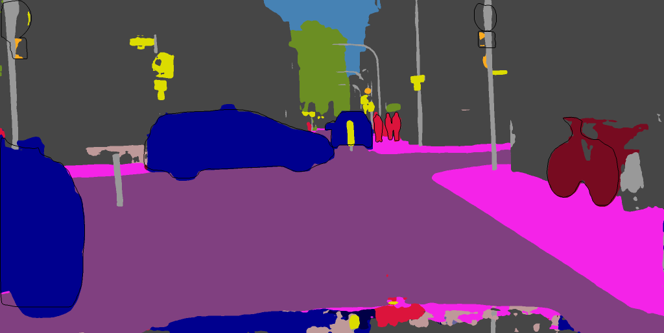}
        \caption{Pseudo-label}
        \label{fig:cityscapes_mask} 
    \end{subfigure}
    \begin{subfigure}[r]{0.32\textwidth}
        \includegraphics[width=\textwidth]{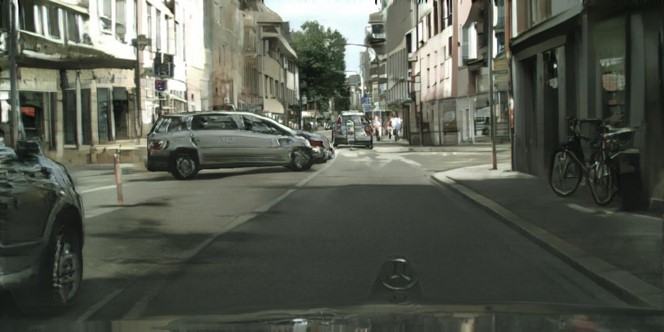}
        \caption{Paired synthetic Cityscapes B image}
        \label{fig:fake_cityscapes}
    \end{subfigure}
\caption{Example from the paired dataset generation. We pseudo-label an unlabeled Cityscapes image (a) with HRNetW48 + OCR + SegFix, resulting in a segmentation mask (b), and vid2vid transforms it to a paired synthetic image (c).}
\label{fig:synthetic_cityscapes}
\end{figure*}

The video-to-video synthesis tool vid2vid generates a video sequence using both a sequence of instance and pixel-level semantic segmentation masks.
Following our approach described in \refsec{sec:approach_datasets}, we generated two different datasets via vid2vid:
\subsubsection{Real-Synthetic Paired Dataset: Cityscapes - synthetic Cityscapes}
\label{sec:experiments_data_gen_paired}
We transform the 500 Cityscapes validation segmentation masks (corresponding to three different video sequences) with fine annotations (containing 30 classes) to three synthetic video sequences via vid2vid in Cityscapes style. We call the resulting video sequences with the corresponding labels \emph{synthetic Cityscapes A}, in contrast to the original Cityscapes video sequences and labels, which we call \emph{real Cityscapes A}. Note that both datasets share the same segmentation labels and differ only in the corresponding image frames.
Vid2vid requires the labeled images to be time-consistent to generate high-quality sequences -- a condition that the aforementioned dataset does not completely fulfill due to its frequent temporal discontinuities.
Thus we follow the suggestion of \cite{vid2vid} and generate pseudo-labels for the unlabeled Cityscapes validation images, which are more time-consistent,  with HRNetW48 + OCR + SegFix	\cite{YuanW18, HuangYGZCW19,YuanCW20,YuanXCW20}\footnote{\url{https://github.com/openseg-group/openseg.pytorch}}  (we use without modifications the variant pretrained on Cityscapes) and Mask Scoring R-CNN for instance segmentation\cite{he2017mask} (pretrained on MS COCO \cite{lin2014microsoft} and further fine-tuned on a proprietary dataset) 
that are also synthesized with the pretrained vid2vid model\footnote{This setup leads to an improved segmentation compared to the HRNetW-48 under test.}. This results in 15,000 new images that, together with the pseudo-labels, we call  \emph{synthetic Cityscapes B} in contrast to the pseudo-labels and original video sequences, which we call \emph{real Cityscapes B}\footnote{Note that the real Cityscapes A images are contained in real Cityscapes B. However, for consistency, we use the pseudo-labels as ground truth here.}. The schematic processing of datasets is depicted in \reffig{fig:experiments_data_generation}.
The transformation process from an unlabeled Cityscapes image (from real Cityscapes B) to its synthetic counterpart (synthetic Cityscapes B) is displayed in the example in \reffig{fig:synthetic_cityscapes}.  Notice the quality of the generated image despite the pseudo-labels.

\subsubsection{Extending input coverage: Transforming CARLA to Cityscapes style}
\begin{figure}[t]
\centering
        \includegraphics[width=0.33\textwidth]{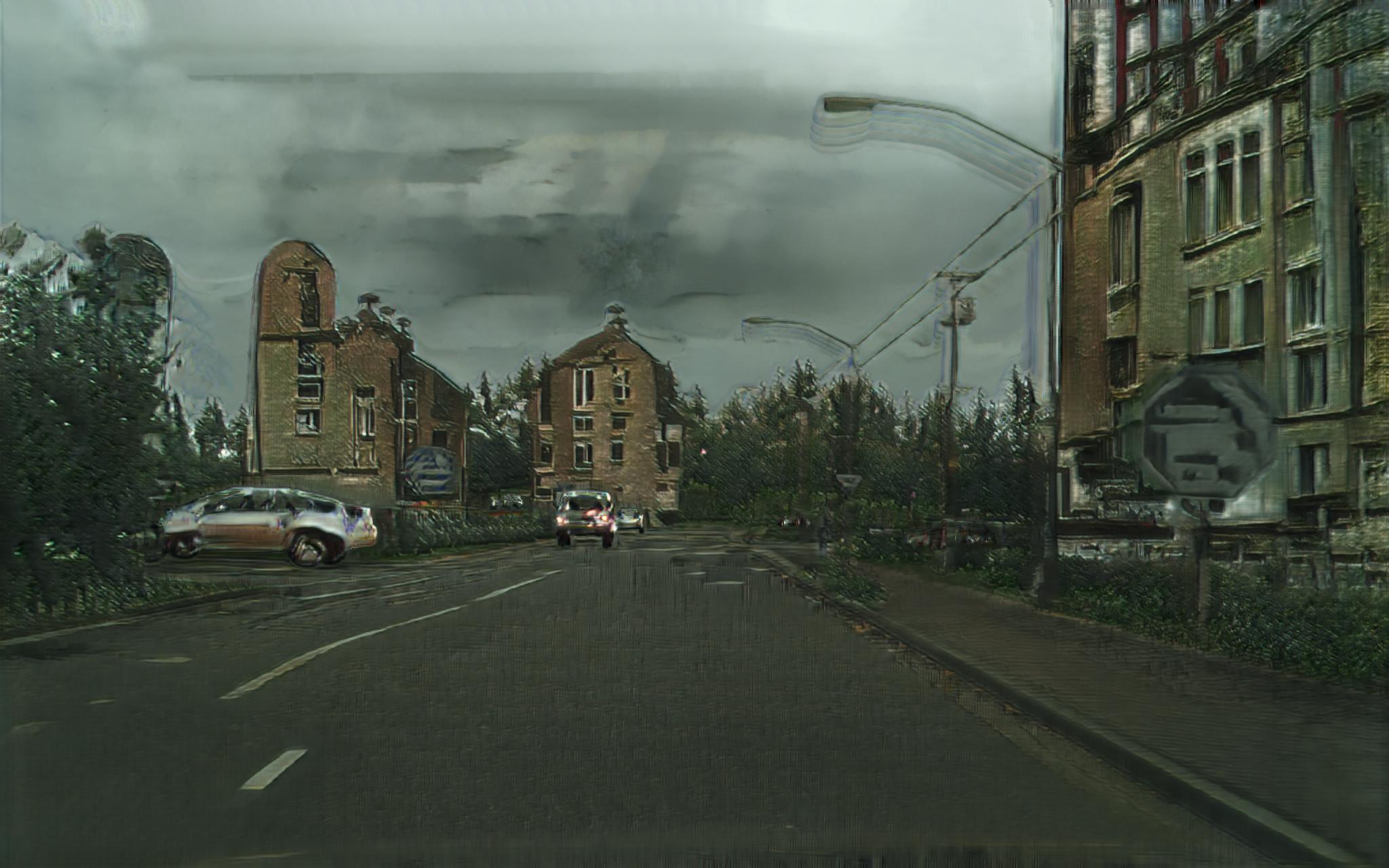}
\caption{Example of a vid2vid-generated image from a CARLA segmentation mask. Notice how the artifact on the right almost has stop sign shape.}
\label{fig:synthetic_carla}
\end{figure}

We generate 100 random video scenes (\ie random city, weather, lighting, etc.) of about 140 images per sequence with a proprietary variant of the CARLA simulator \cite{CARLA2017} together with labels, calling this \emph{original CARLA} dataset. The corresponding segmentation masks are mapped to the 30 Cityscapes classes and processed by vid2vid.
This, together with the labels, now constitutes the \emph{transformed CARLA} dataset. For the schematic process see \reffig{fig:experiments_data_generation}. An example image from transformed CARLA is displayed in \reffig{fig:synthetic_carla}. 

\subsection{Validation of transferability}

\begin{figure}[h]
\centering
    \begin{subfigure}[b]{0.49\textwidth}
        \includegraphics[width=\textwidth]{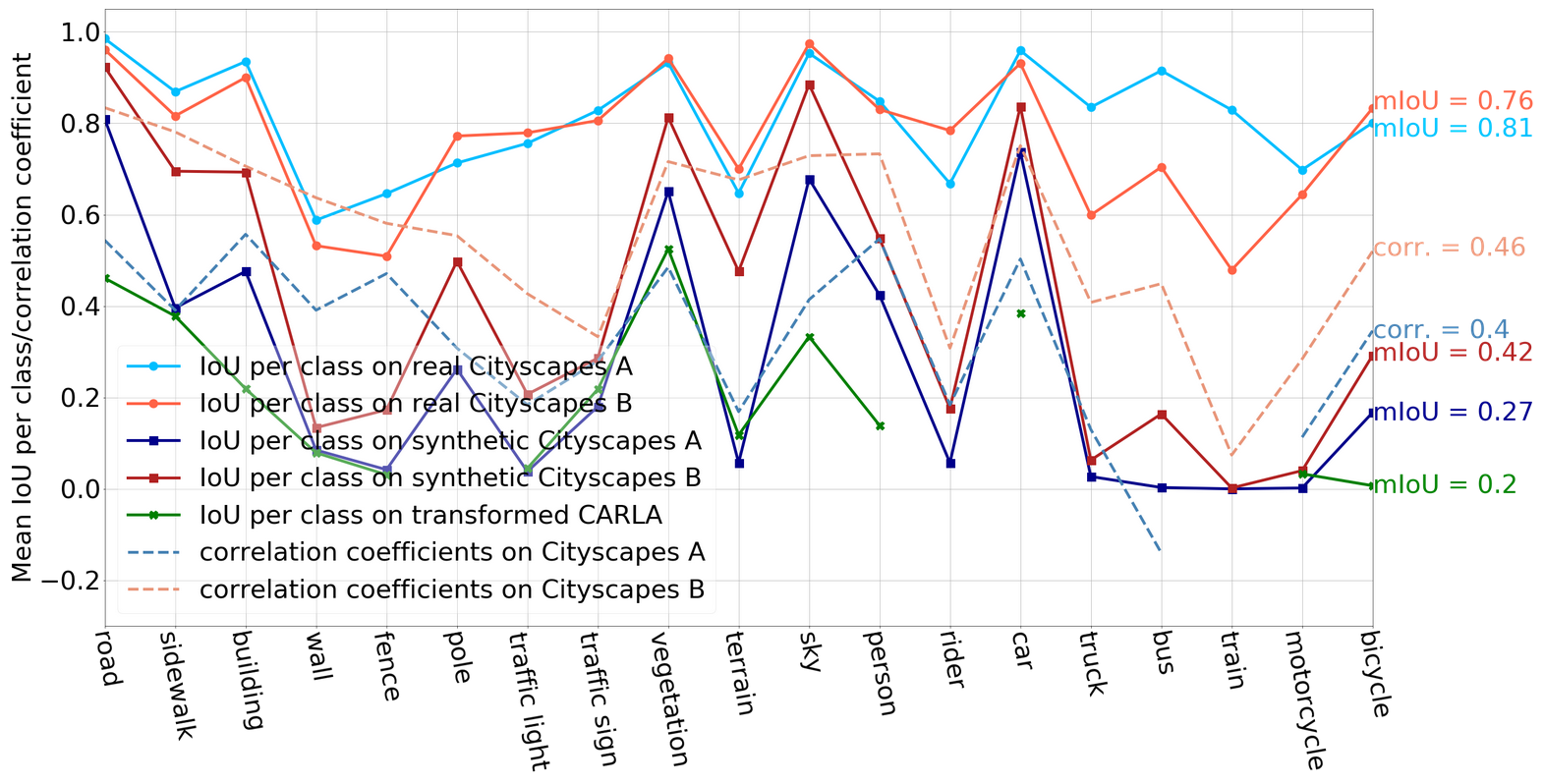}
        \label{fig:Cityscapes_and_seq_Cityscapes_miou}
    \end{subfigure}
\caption{Correlation and performance analysis results on paired Cityscapes A and B as well as transformed CARLA classwise IoUs. Note that there is no correlation coefficient for class 'train' on Cityscapes A since there are not enough samples from the synthetic and real sets. Classes in CARLA that have no IoU value are not available from simulation.} 
\label{fig:correlation_analysis}
\end{figure}

\subsubsection{Correlation and performance analysis}
\begin{figure}[t]
\centering
        \begin{subfigure}[b]{0.49\textwidth}
            \includegraphics[width=\textwidth]{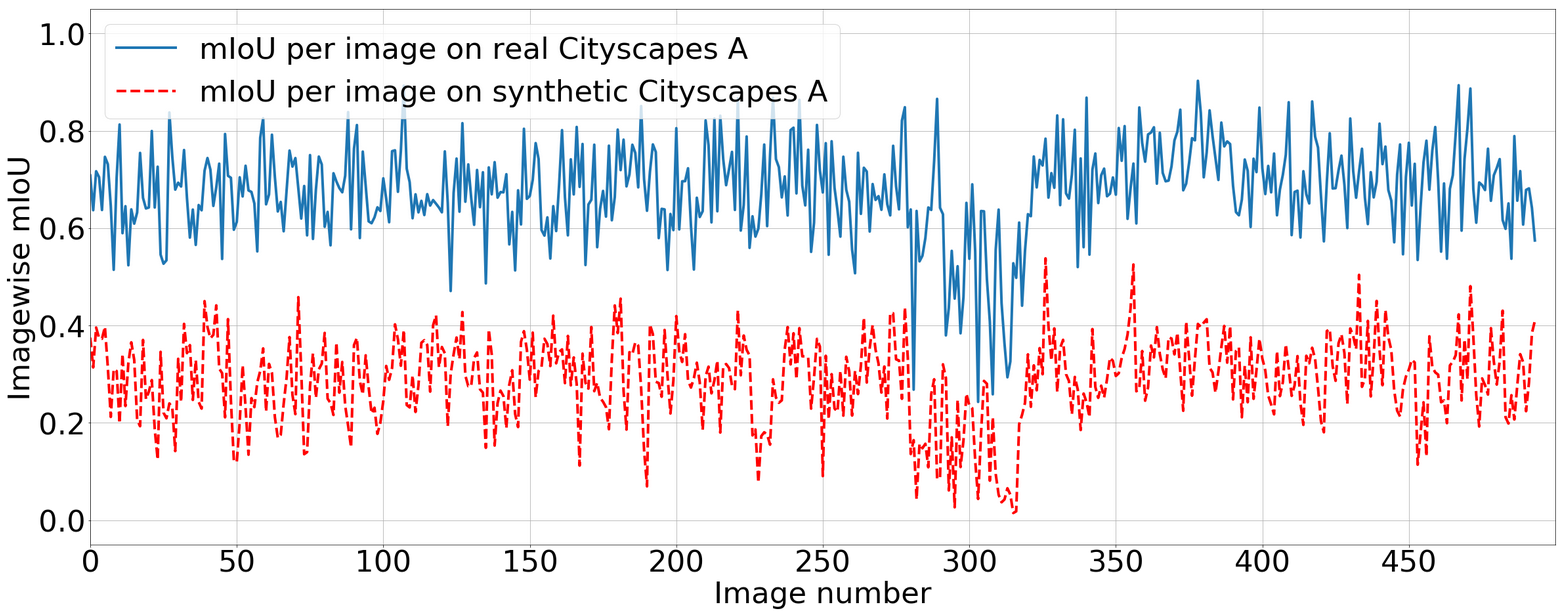}
            \includegraphics[width=\textwidth]{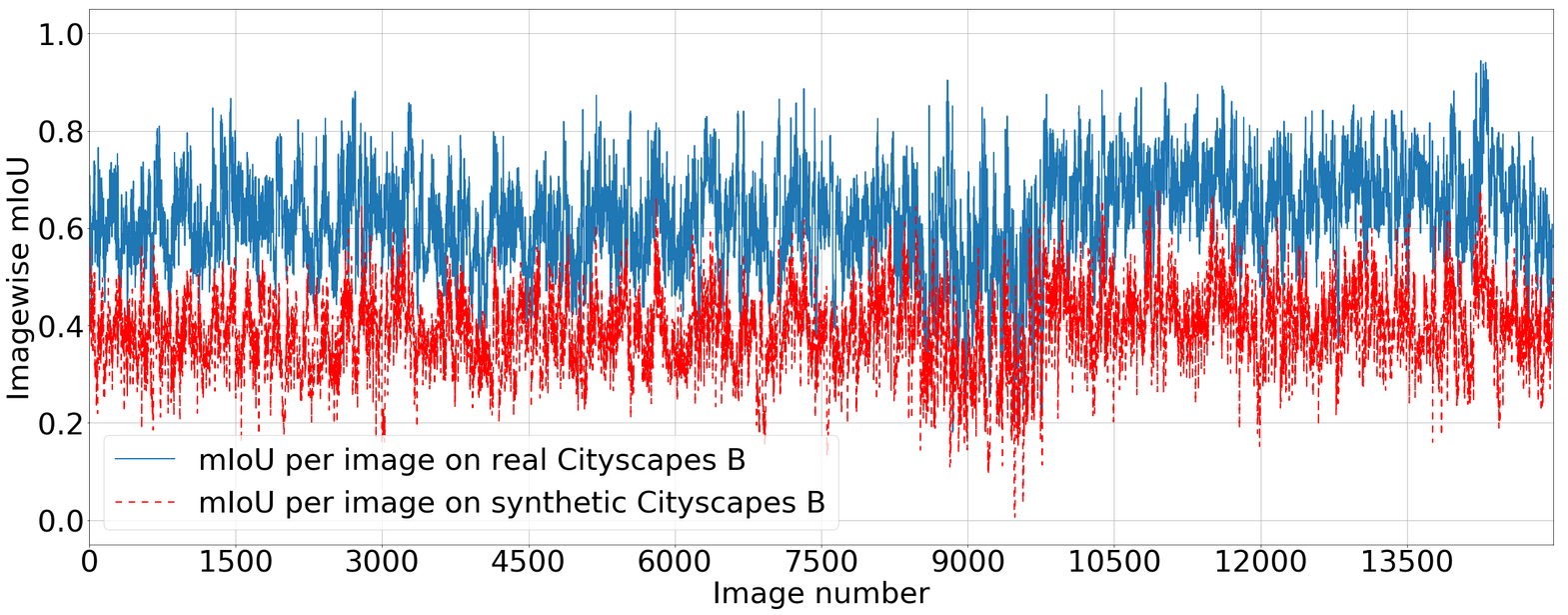}
        \end{subfigure}
        \caption{Correlation of mIoU per image on paired Cityscapes A (top) and on paired Cityscapes B (bottom).} 
        \label{fig:correlation_Cityscapes}
\label{fig:correlation_analysis_miou}
\end{figure}

We conduct the performance analysis on the paired set, synthetic and real Cityscapes A, described 
in \refsec{sec:experiments_data_gen_paired} using mIoU as the performance metric. We use the reduced set of $19$ Cityscapes classes and distinguish between them to gather IoUs per class as well as mIoU scores per image. 
We average these quantities over all images of the respective dataset and compute the sample correlation coefficients as described in \refsec{sec:approach_validation_correlation}. We repeat this procedure for paired Cityscapes B. The obtained results can be seen 
in \reffig{fig:correlation_analysis} and \reffig{fig:correlation_analysis_miou}.
\reffig{fig:correlation_analysis}
depicts class-wise IoU and class-wise correlation coefficients on both paired sets, in which one can observe a relatively high class-wise correlation coefficient when the network performs well  (\ie it has a rather good IoU for that class), \eg for the classes road, building and vegetation as well as the particularly safety-relevant classes person and car. Note that despite a high IoU for both synthetic and real sets for the class sky, the correlation coefficient is relatively low.

Plotting the mIoU on paired real and synthetic Cityscapes A in the upper part of \reffig{fig:correlation_Cityscapes} shows how they correlate (the sample correlation coefficient is approximately 0.403). The lower part of \reffig{fig:correlation_Cityscapes} shows an analogous plot for the B sets, in which the sample correlation coefficient is slightly higher with 0.457.
The displayed peaks and dips of the image-wise mIoU scores on the synthetic set matching those on the real set qualitatively suggest a rather good transferability. From a testing perspective, especially the negative correspondence, \ie low performance on synthetic data coinciding with lower real data performance, are important as they might help reveal failure modes.
Evaluation of the function on the extended scenes of transformed CARLA yields a mIoU of 0.196, constituting a decrease of around 0.07 compared to synthetic Cityscapes A. However, the performance on the transformed CARLA seems to depend on the choice of the particular sequence of videos.
This might provide a first insight that semantic concepts in these sequences might constitute failure modes.

\subsubsection{Error distribution analysis}
\begin{figure}[h]
\centering
\begin{subfigure}[b]{0.49\textwidth}
    \begin{subfigure}[b]{\textwidth}
        \includegraphics[width=\textwidth]{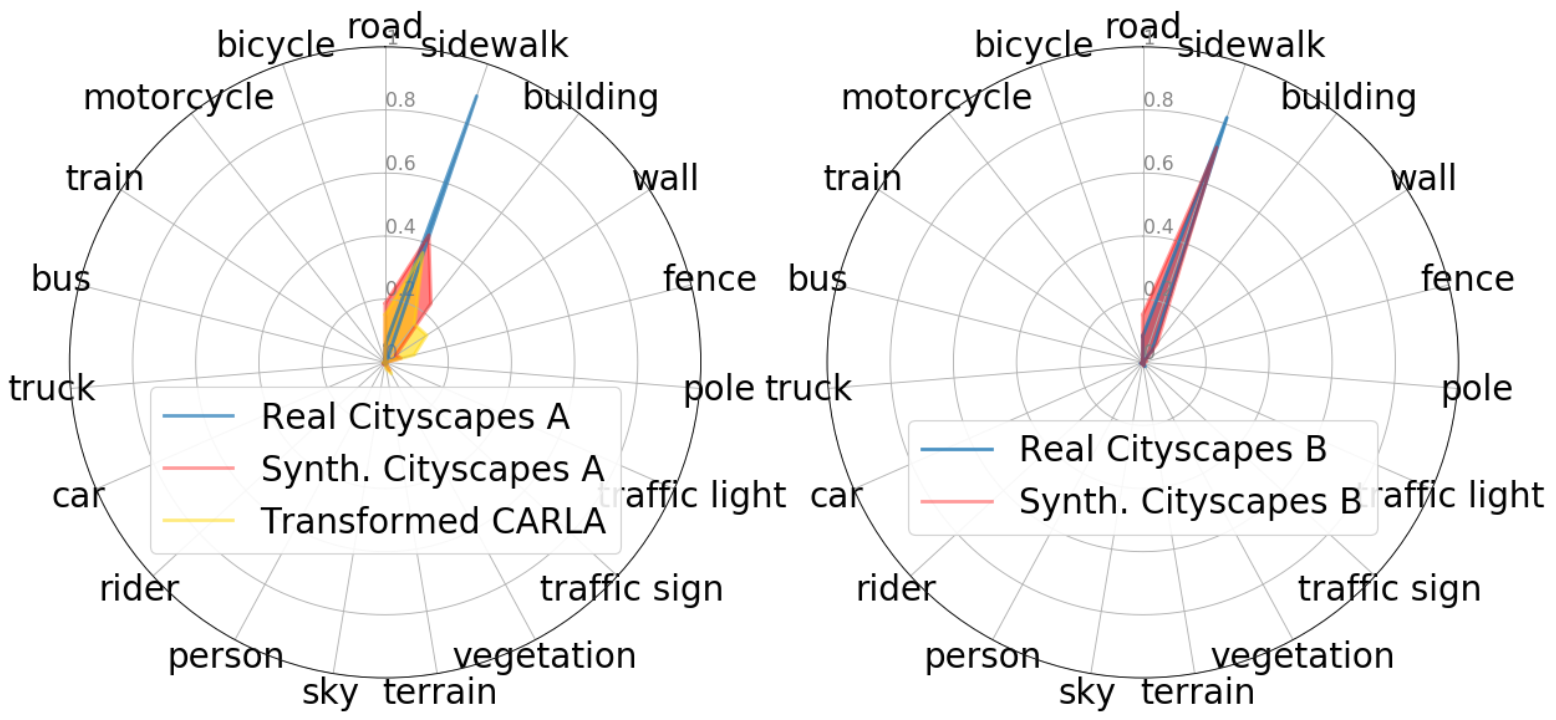}
    \end{subfigure}
\end{subfigure}
\caption{Error distribution analysis: Radar plots \wrt class sidewalk for paired Cityscapes A and transformed CARLA (left) as well as on paired Cityscapes B (right).}
        \label{fig:radar plot1}
\label{fig:radar_plots}
\end{figure}

Conducting the error distribution analysis described in \refsec{sec:approach_validation_radarplots}, we find that $\mathrm{FNS_{c,k}}$ values differ for the datasets, \ie the error distribution across the datasets differs, providing evidence that errors are not always transferable (in this work we only show one example radar plot, omitting the rest). This effect is in particular visible for transformed CARLA errors.
On paired Cityscapes B, we see that errors are rather comparable,
especially for the larger classes such as road, sidewalk, building, wall, and sky. We observe that the better quality of the real B set, relative to the requirements of vid2vid, enhances the comparability of errors. However, the outcome is still class-dependent, hinting at the fact that the current simulated data might only be suitable for testing \wrt some particular classes. 
We observe that errors tend to fall into naturally adjacent classes. The example radar plot in \reffig{fig:radar_plots} shows the ground truth class sidewalk getting mistaken for road and building. By definition our $\text{FNS}_{c,k}$ measures the (relative) amount of wrongly segmented pixel area.  
Similar to IoU, we expect $\text{FNS}_{c,k}$ to be more fluctuating for objects of smaller area, which is apparent in the larger stability for aforementioned classes with typically large pixel areas. Lastly, we expect the measure to be correlated to the IoU itself, as they are related quantities: To be precise, the smaller the IoU of an object, the larger it contributes to $\text{FNS}_{c,k}$. 

\subsubsection{Discriminator on model outputs and errors}

\begin{figure*}[t]
\centering
    \begin{subfigure}[b]{0.32\textwidth}
        \includegraphics[width=\textwidth]{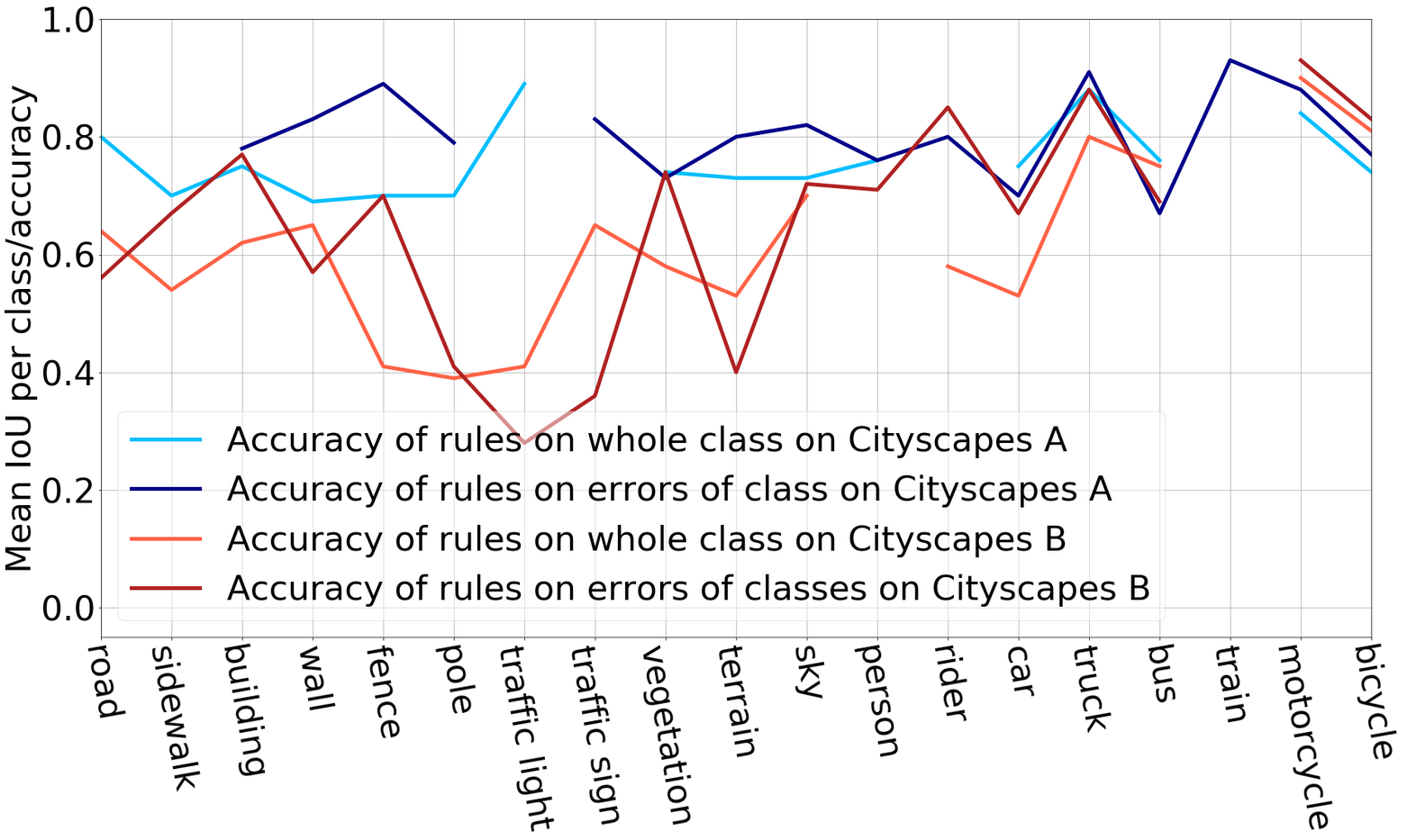}
        \caption{Classwise accuracy scores of rule-sets on model outputs and errors on paired Cityscapes A and B.}
        \label{fig:acc_rules}
    \end{subfigure}
    \begin{subfigure}[b]{0.32\textwidth}
        \includegraphics[width=\textwidth]{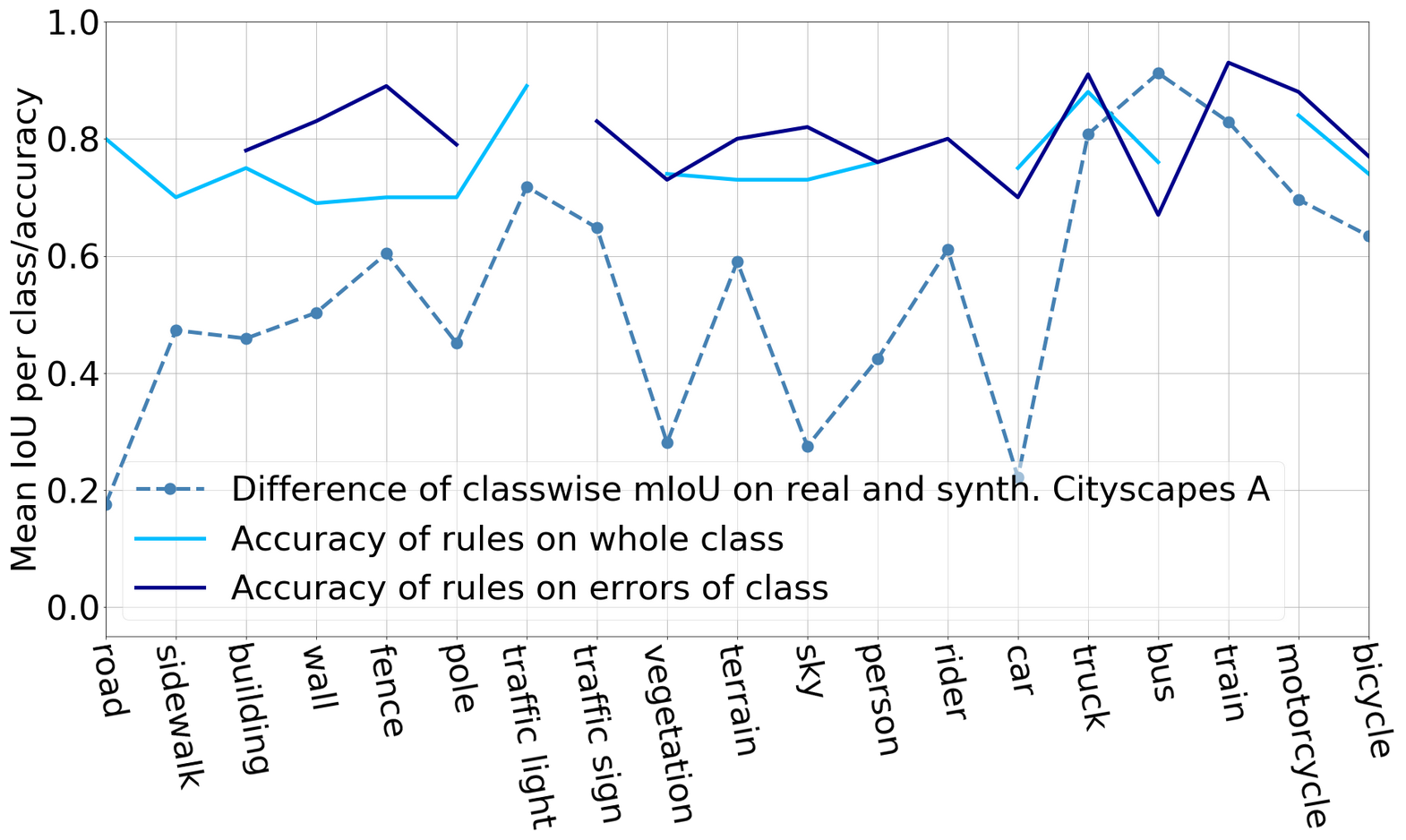}
        \caption{Classwise rule-set accuracy scores together with differences in IoU performance on paired Cityscapes A.}
        \label{fig:acc_miou_city} 
    \end{subfigure}
    \begin{subfigure}[b]{0.32\textwidth}
        \includegraphics[width=\textwidth]{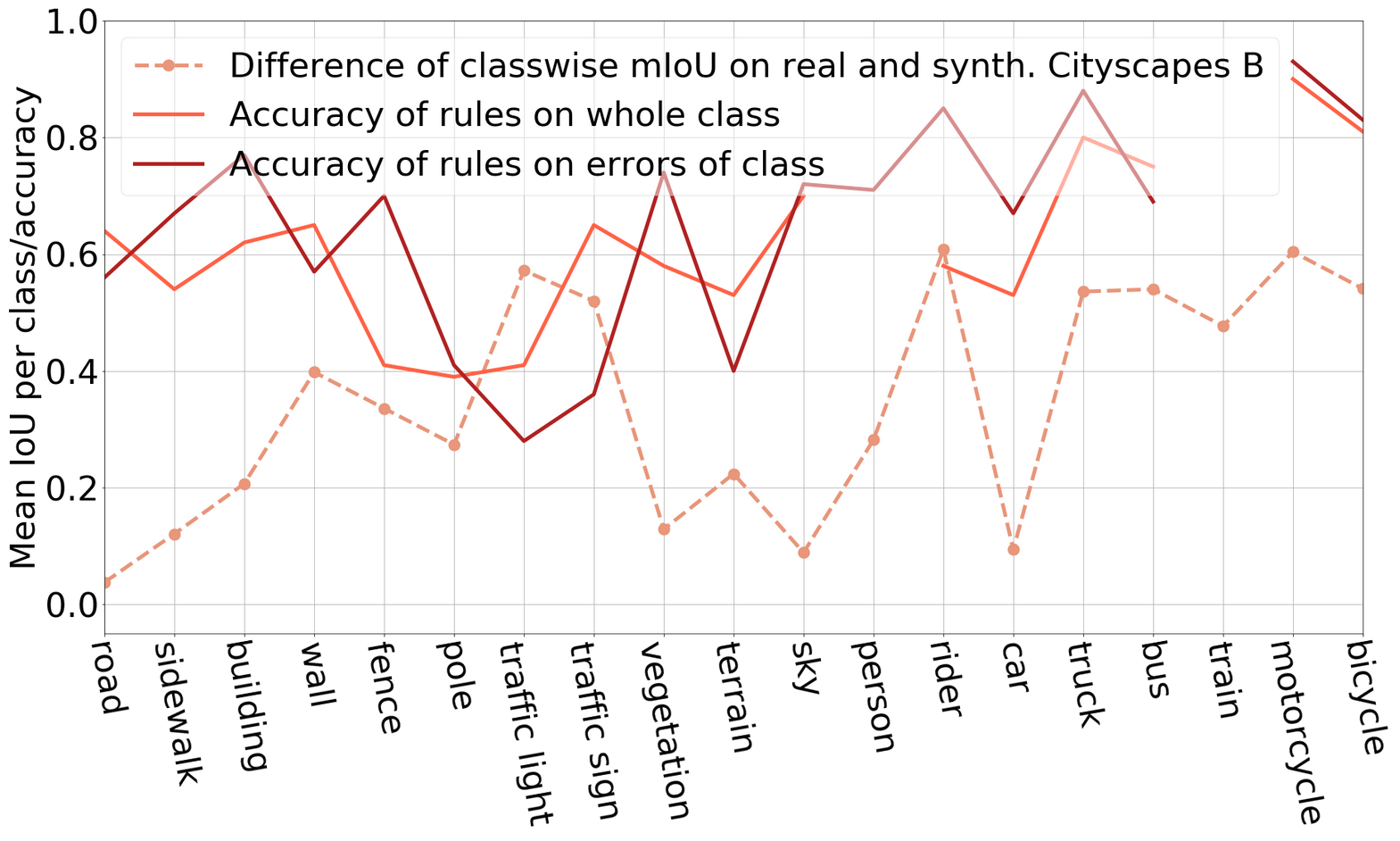}
        \caption{Classwise rule-set accuracy scores together with differences in IoU performance on paired Cityscapes B.}
        \label{fig:acc_miou_seq_city}
    \end{subfigure}
\caption{Results from rule learning on model outputs and errors on paired Cityscapes A and B. Note that for some classes no rules could be found or that there were not enough components to learn rules, leading  to lacking data points in the plots.}
\label{fig:discriminator}
\end{figure*}

We train the SkopeRules rule learner based on feature engineering of MetaSeg, \cite{MetaSeg}, to distinguish 
whether a model output/error belongs to a real or synthetic input image. 
More precisely, using \cite{MetaSeg}, we compute various quantities from the predicted pixel-wise class probabilities, which are aggregated per connected component (segment) of the model's output segmentation mask. These quantities include dispersion measures, \ie the pixel-wise entropy, probability margin, and variation ratio. These measures get aggregated over each whole predicted segment as well as only the corresponding boundary and the inner. The aggregation is performed by considering both mean and variance over all pixel-values that correspond to the whole segment, the inner or the boundary, respectively. Furthermore, we consider the size measures, \ie the size of the whole segment, its boundary and its inner as well as fractality measures like the segment size over the boundary size. Altogether we obtain a structured dataset that contains a number of interpretable scalar values per predicted segment. For further details, see \cite{MetaSeg}.
Moreover, MetaSeg stores labels and IoUs for each segment. This diversity of computed metrics allows for a distinct uncertainty assessment for the predicted segments, enabling a geometric interpretation of the sets of rules. For instance, a rule for some class $c$ including boundary entropy as a classifier for real vs.\ synthetic input implies that differences between the sets lie on the boundaries of the respective class segments. 
\par
In total, we compute $35$ different metrics per segment of the prediction mask of the HRNet for each image of paired Cityscapes A and B, respectively, and save information about the belonging dataset. 
We then separate the dataset according to the $19$ semantic classes of Cityscapes (classwise choosing the minority dataset - real or synthetic - as the target to learn rules for)  
and perform a random $80:20$ train-test split.
In a first step, we learn rules on all classwise segments before we filter (again classwise) for errors, \ie for segments with $\text{IoU} =0$. 
The same analysis is performed on a subset (of the same size as Cityscapes A) of images from paired Cityscapes B. 
The accuracy scores of the resulting rule sets can be seen in \reffig{fig:discriminator}.\footnote{We used the (top $k$) rules optimized for the minority class, thus sometimes degrading overall accuracy.}
Overall, the accuracy scores are rather high (see \reffig{fig:acc_rules}) -- our discriminator can (easily) tell the inputs apart -- implying that the model behavior differs on real and synthetic paired input data. Additionally, on both Cityscapes A and B, the restriction to only error components leads to higher mean accuracy of the set of rules \ie a better distinguishability of real and synthetic input data. However, as seen in \reffig{fig:acc_rules}, the accuracy scores on Cityscapes B are lower on average, indicating that better quality of synthetic data makes it more difficult to differentiate between real and synthetic inputs. 
We observe that the rule accuracy scores do not reflect any IoU performance gap, as we see in \reffig{fig:acc_miou_city} and \reffig{fig:acc_miou_seq_city}. This underlines our claim that performance metrics such as (m)IoU (as described in \refsec{sec:approach_validation_correlation}) alone cannot assess the comparability of model behavior.
Interestingly, concerning hyperparameters, a maximal depth of $1$ turns out to be optimal for the rule sets across all classes in all our discriminator experiments. 
Also, the rule sets for model outputs on paired Cityscapes A contain mostly boundary metrics, whereas the differences between the datasets are rather scattered and thus more difficult to interpret for the remaining experiments. This might be due to the increased difficulty of synthesizing boundary pixels via vid2vid.
\par Finally, we can say that our proposed methods and metric correlate well with the visually perceptible quality difference in the synthetic set: On the more realistic looking synthetic Cityscapes B, rule accuracy scores drop. 
Nevertheless, the results of the discriminator analysis 
show that the proposed metrics provide a good way to identify the domain shift present between real and synthesized datasets. Note, however, that for practical testing purposes it may not be necessary to fully close this gap.

%% file: 5_conclusion.tex
\section{Conclusion and Discussion}
\label{sec:conclusion}
We presented a conceptual framework to validate simulation-based testing of real-world ML applications, which we instantiated on a semantic segmentation task in the context of AD.
As simulation and real-world data have a domain gap, our work explicitly addresses the question of \textit{transferability} of testing results.
We employ a generative label-to-image transfer method, mapping from the ground truth labels back to the (real world) image domain, which provides two key advantages:
First, we can map a given labeled (real) dataset onto a synthesized version of itself allowing us to directly and in detail investigate the resulting domain gap incurred by the generative method.
Second, applying the same generative model to ground truth data from any source, \eg a simulator, we can test the ML application.
Under the condition that the simulated ground truth is of the same form as the real world one, we have almost no domain gap between the synthesized real data and the data synthesized from a simulation.
So, using this two-stage approach we can largely bypass the question of domain gap regarding simulated test data, and instead shift it to a more controllable comparison between two datasets with identical ground truths.

While the performance of the generator is not crucial, it is clear that our approach still benefits from a small domain gap between the actual and the synthesized data.
Improvements regarding better generative models, more available data, and additional validation metrics can easily be incorporated into our modular framework.
With semantically rich enough labels to facilitate data generation, 
our approach could be used on other tasks such as \eg object detection, using mean average precision as performance score, and on other application domains as,
\eg in the context of text mining.

Turning, at last, to the concrete instantiating of the framework on the segmentation task, we evaluated transferability calculating class-wise mIoU correlation coefficients and found for cars or person surprisingly strong and encouraging values of $0.7$.
A deeper analysis of failure modes based on manual feature extraction, however, revealed that failures can be still clearly classified as belonging to the real data or its synthesized counterpart.
Lastly, while we demonstrated the feasibility of the approach the actual test of the segmentation model, \eg active weak-spot search, is left for future work.

%% file: main.bbl
% Generated by IEEEtran.bst, version: 1.14 (2015/08/26)
\begin{thebibliography}{10}
\providecommand{\url}[1]{#1}
\csname url@samestyle\endcsname
\providecommand{\newblock}{\relax}
\providecommand{\bibinfo}[2]{#2}
\providecommand{\BIBentrySTDinterwordspacing}{\spaceskip=0pt\relax}
\providecommand{\BIBentryALTinterwordstretchfactor}{4}
\providecommand{\BIBentryALTinterwordspacing}{\spaceskip=\fontdimen2\font plus
\BIBentryALTinterwordstretchfactor\fontdimen3\font minus
  \fontdimen4\font\relax}
\providecommand{\BIBforeignlanguage}[2]{{%
\expandafter\ifx\csname l@#1\endcsname\relax
\typeout{** WARNING: IEEEtran.bst: No hyphenation pattern has been}%
\typeout{** loaded for the language `#1'. Using the pattern for}%
\typeout{** the default language instead.}%
\else
\language=\csname l@#1\endcsname
\fi
#2}}
\providecommand{\BIBdecl}{\relax}
\BIBdecl

\bibitem{BRAIEK2020110542}
H.~B. Braiek and F.~Khomh, ``On testing machine learning programs,''
  \emph{Journal of Systems and Software}, vol. 164, p. 110542, 2020.

\bibitem{MLtestingsurvey2020}
J.~M. {Zhang}, M.~{Harman}, L.~{Ma}, and Y.~{Liu}, ``{Machine Learning Testing:
  Survey, Landscapes and Horizons},'' \emph{Transactions on Software
  Engineering}, 2020.

\bibitem{simulation_based_testing20}
I.~{Paranjape}, A.~{Jawad}, Y.~{Xu}, A.~{Song}, and J.~{Whitehead}, ``{A
  Modular Architecture for Procedural Generation of Towns, Intersections and
  Scenarios for Testing Autonomous Vehicles},'' in \emph{Intelligent Vehicles
  Symposium (IV)}, 2020, pp. 162--168.

\bibitem{wagner19}
S.~{Wagner}, K.~{Groh}, T.~{Kühbeck}, and A.~{Knoll}, ``{Towards
  Cross-Verification and Use of Simulation in the Assessment of Automated
  Driving},'' in \emph{Intelligent Vehicles Symposium (IV)}, 2019, pp.
  1589--1596.

\bibitem{CARLA2017}
A.~Dosovitskiy, G.~Ros, F.~Codevilla, A.~Lopez, and V.~Koltun, ``{CARLA}: {An}
  open urban driving simulator,'' in \emph{1st Annual Conference on Robot
  Learning}, 2017, pp. 1--16.

\bibitem{vonrueden2020informed}
L.~von Rueden, S.~Mayer, K.~Beckh, B.~Georgiev, S.~Giesselbach, R.~Heese,
  B.~Kirsch, J.~Pfrommer, A.~Pick, R.~Ramamurthy, M.~Walczak, J.~Garcke,
  C.~Bauckhage, and J.~Schuecker, ``{Informed Machine Learning - A Taxonomy and
  Survey of Integrating Knowledge into Learning Systems},'' \emph{arXiv
  preprint arXiv:1903.12394v2}, 2020.

\bibitem{von2020combining}
L.~von Rueden, S.~Mayer, R.~Sifa, C.~Bauckhage, and J.~Garcke, ``Combining
  machine learning and simulation to a hybrid modelling approach: Current and
  future directions,'' in \emph{International Symposium on Intelligent Data
  Analysis}.\hskip 1em plus 0.5em minus 0.4em\relax Springer, 2020, pp.
  548--560.

\bibitem{online_offline_testing}
F.~U. {Haq}, D.~{Shin}, S.~{Nejati}, and L.~C. {Briand}, ``{Comparing Offline
  and Online Testing of Deep Neural Networks: An Autonomous Car Case Study},''
  in \emph{International Conference on Software Testing, Validation and
  Verification (ICST)}, 2020, pp. 85--95.

\bibitem{safety_args2018}
A.~Rudolph, S.~Voget, and J.~Mottok, ``{A Consistent Safety Case Argumentation
  for Artificial Intelligence in Safety Related Automotive Systems},'' in
  \emph{{European Congress on Embedded Real Time Software and Systems (ERTS)}},
  2018.

\bibitem{safety_args_2020_schwalbe}
G.~Schwalbe and M.~Schels, ``{A Survey on Methods for the Safety Assurance of
  Machine Learning Based Systems},'' in \emph{{European Congress on Embedded
  Real Time Software and Systems (ERTS)}}, 2020.

\bibitem{barbier2019validation}
M.~{Barbier}, A.~{Renzaglia}, J.~{Quilbeuf}, L.~{Rummelhard}, A.~{Paigwar},
  C.~{Laugier}, A.~{Legay}, J.~{Ibañez-Guzmán}, and O.~{Simonin},
  ``{Validation of Perception and Decision-Making Systems for Autonomous
  Driving via Statistical Model Checking},'' in \emph{Intelligent Vehicles
  Symposium (IV)}, 2019, pp. 252--259.

\bibitem{Neurohr20}
C.~{Neurohr}, L.~{Westhofen}, T.~{Henning}, T.~{de Graaff}, E.~{Möhlmann}, and
  E.~{Böde}, ``Fundamental considerations around scenario-based testing for
  automated driving,'' in \emph{Intelligent Vehicles Symposium (IV)}, 2020, pp.
  121--127.

\bibitem{Bussler20}
A.~{Bussler}, L.~{Hartjen}, R.~{Philipp}, and F.~{Schuldt}, ``{Application of
  Evolutionary Algorithms and Criticality Metrics for the Verification and
  Validation of Automated Driving Systems at Urban Intersections},'' in
  \emph{Intelligent Vehicles Symposium (IV)}, 2020, pp. 128--135.

\bibitem{kouw2019review}
W.~M. Kouw and M.~Loog, ``{A Review of Domain Adaptation without Target
  Labels},'' \emph{arXiv preprint arXiv:1901.05335}, 2019.

\bibitem{Mei2018}
W.~Mei and W.~Deng, ``{Deep Visual Domain Adaptation: A Survey},''
  \emph{Neurocomputing}, 2018.

\bibitem{Chen_2017_ICCV}
D.~Chen, J.~Liao, L.~Yuan, N.~Yu, and G.~Hua, ``Coherent online video style
  transfer,'' in \emph{International Conference on Computer Vision (ICCV)},
  2017.

\bibitem{vid2vid}
T.-C. Wang, M.-Y. Liu, J.-Y. Zhu, G.~Liu, A.~Tao, J.~Kautz, and B.~Catanzaro,
  ``{Video-to-Video Synthesis},'' in \emph{Advances in Neural Information
  Processing Systems}.\hskip 1em plus 0.5em minus 0.4em\relax Curran
  Associates, Inc., 2018, pp. 1144--1156.

\bibitem{Isola_2017_CVPR}
P.~Isola, J.-Y. Zhu, T.~Zhou, and A.~A. Efros, ``{Image-To-Image Translation
  With Conditional Adversarial Networks},'' in \emph{Conference on Computer
  Vision and Pattern Recognition (CVPR)}, 2017.

\bibitem{pix2pix_follow_up}
T.~{Wang}, M.~{Liu}, J.~{Zhu}, A.~{Tao}, J.~{Kautz}, and B.~{Catanzaro},
  ``{High-Resolution Image Synthesis and Semantic Manipulation with Conditional
  GANs},'' in \emph{Conference on Computer Vision and Pattern Recognition
  (CVPR)}, 2018, pp. 8798--8807.

\bibitem{Meta-Sim}
A.~Kar, A.~Prakash, M.-Y. Liu, E.~Cameracci, J.~Yuan, M.~Rusiniak, D.~Acuna,
  A.~Torralba, and S.~Fidler, ``{Meta-Sim: Learning to Generate Synthetic
  Datasets},'' in \emph{International Conference on Computer Vision (ICCV)},
  2019.

\bibitem{Meta-Sim2}
J.~Devaranjan, A.~Kar, and S.~Fidler, ``Meta-sim2: Unsupervised learning of
  scene structure for synthetic data generation,'' in \emph{European Conference
  on Computer Vision (ECCV)}, 2020, pp. 715--733.

\bibitem{Shrivastava_2017_CVPR}
A.~Shrivastava, T.~Pfister, O.~Tuzel, J.~Susskind, W.~Wang, and R.~Webb,
  ``{Learning From Simulated and Unsupervised Images Through Adversarial
  Training},'' in \emph{Conference on Computer Vision and Pattern Recognition
  (CVPR)}, 2017.

\bibitem{Virtual_Kitty}
A.~Gaidon, Q.~Wang, Y.~Cabon, and E.~Vig, ``{Virtual Worlds as Proxy for
  Multi-Object Tracking Analysis},'' in \emph{Conference on Computer Vision and
  Pattern Recognition (CVPR)}, 2016.

\bibitem{Kitty}
A.~{Geiger}, P.~{Lenz}, and R.~{Urtasun}, ``{Are we Ready for Autonomous
  Driving? The KITTI Vision Benchmark Suite},'' in \emph{Conference on Computer
  Vision and Pattern Recognition (CVPR)}, 2012, pp. 3354--3361.

\bibitem{HRNet2020}
J.~{Wang}, K.~{Sun}, T.~{Cheng}, B.~{Jiang}, C.~{Deng}, Y.~{Zhao}, D.~{Liu},
  Y.~{Mu}, M.~{Tan}, X.~{Wang}, W.~{Liu}, and B.~{Xiao}, ``{Deep
  High-Resolution Representation Learning for Visual Recognition},''
  \emph{Transactions on Pattern Analysis and Machine Intelligence}, pp. 1--1,
  2020.

\bibitem{Cordts2016Cityscapes}
M.~Cordts, M.~Omran, S.~Ramos, T.~Rehfeld, M.~Enzweiler, R.~Benenson,
  U.~Franke, S.~Roth, and B.~Schiele, ``{The Cityscapes Dataset for Semantic
  Urban Scene Understanding},'' in \emph{Conference on Computer Vision and
  Pattern Recognition (CVPR)}, 2016.

\bibitem{MetaSeg}
M.~{Rottmann}, P.~{Colling}, T.~{Paul Hack}, R.~{Chan}, F.~{Hüger},
  P.~{Schlicht}, and H.~{Gottschalk}, ``{Prediction Error Meta Classification
  in Semantic Segmentation: Detection via Aggregated Dispersion Measures of
  Softmax Probabilities},'' in \emph{International Joint Conference on Neural
  Networks (IJCNN)}, 2020, pp. 1--9.

\bibitem{YuanW18}
Y.~Yuan and J.~Wang, ``{Ocnet: Object Context Network for Scene Parsing},''
  \emph{arXiv preprint arXiv:1809.00916}, 2018.

\bibitem{HuangYGZCW19}
L.~Huang, Y.~Yuan, J.~Guo, C.~Zhang, X.~Chen, and J.~Wang, ``{Interlaced Sparse
  Self-Attention for Semantic Segmentation},'' \emph{arXiv preprint
  arXiv:1907.12273}, 2019.

\bibitem{YuanCW20}
Y.~Yuan, X.~Chen, and J.~Wang, ``{Object-Contextual Representations for
  Semantic Segmentation},'' \emph{arXiv preprint arXiv:1909.11065}, 2020.

\bibitem{YuanXCW20}
Y.~Yuan, J.~Xie, X.~Chen, and J.~Wang, ``Segfix: Model-agnostic boundary
  refinement for segmentation,'' \emph{arXiv preprint arXiv:2007.04269}, 2020.

\bibitem{he2017mask}
K.~He, G.~Gkioxari, P.~Doll{\'a}r, and R.~Girshick, ``{Mask R-CNN},'' in
  \emph{International Conference on Computer Vision (ICCV)}, 2017, pp.
  2961--2969.

\bibitem{lin2014microsoft}
T.-Y. Lin, M.~Maire, S.~Belongie, J.~Hays, P.~Perona, D.~Ramanan,
  P.~Doll{\'a}r, and C.~L. Zitnick, ``{Microsoft {COCO}: Common objects in
  context},'' in \emph{European Conference on Computer Vision (ECCV)}, 2014,
  pp. 740--755.

\end{thebibliography}
